\documentclass[letterpaper, 10 pt, conference]{ieeeconf}  

\IEEEoverridecommandlockouts                              

\usepackage{xspace}
\usepackage{hyperref}
\usepackage{booktabs} 
\usepackage{tabularx}
\usepackage{multirow}
\usepackage{xcolor}
\usepackage{graphicx}
\usepackage{pifont}
\usepackage[utf8]{inputenc}
\usepackage{amsmath}
\usepackage{amssymb}
\usepackage{stmaryrd}
\usepackage{subcaption}
\usepackage{url}
\usepackage{listings}
\usepackage{caption}
\usepackage{algorithm, algorithmic}

\newcommand{\BibTeX}{\rm B\kern-.05em{\sc i\kern-.025em b}\kern-.08em\TeX}
\newcommand{\adexgym}{MATS-Gym\xspace}
\newcommand{\cmark}{\ding{51}}%
\newcommand{\xmark}{\ding{55}}%

\newcommand{\counterfactual}[1]{\ensuremath{%
  \ {\Box}_{#1}\kern-6pt
    \raise1pt\hbox{$\mathord{\longrightarrow}$\ }}}

\title{\LARGE \bf
Scenario-Based Curriculum Generation for Multi-Agent Driving
}

\author{Axel Brunnbauer$^{1}$, Luigi Berducci$^{1}$, Peter Priller$^{2}$, Dejan Nickovic$^{3}$, Radu Grosu$^{1}$
\thanks{$^{1}$ CPS, Technische Universit\"at Wien (TU Wien), Austria}%
\thanks{$^{2}$ AVL List GmbH}%
\thanks{$^{3}$ Austrian Institute of Technology, AIT}%
 \thanks{$^{a}$ Correspondence: {\tt\small axel.brunnbauer@tuwien.ac.at}}%
}

\begin{document}

\maketitle
\thispagestyle{empty}
\pagestyle{empty}

\begin{abstract}

The automated generation of diversified training sce\-na\-ri\-os has been an important ingredient in many complex learning tasks, especially in real-world application domains such as autonomous driving, where auto-curriculum generation is considered vital for obtaining robust and general policies. However, crafting traffic scenarios with multiple, heterogeneous agents, is typically considered as a tedious and time-consuming task, especially in more complex simulation environments. 
To this end, we introduce \adexgym, a multi-agent training framework for autonomous driving, that uses partial-scenario specifications to generate traffic scenarios with a variable number of agents which are executed in CARLA, a high-fidelity driving simulator. 
\adexgym reconciles scenario execution engines, such as Scenic and ScenarioRunner, with established multi-agent training frameworks where the interaction between the environment and the agents is modeled as a partially-observable stochastic game. 
Furthermore, we integrate \adexgym with techniques from unsupervised environment design to automate the generation of adaptive auto-curricula, which is the first application of such algorithms to the domain of autonomous driving.
The code is available at \url{https://github.com/AutonomousDrivingExaminer/mats-gym}.


\end{abstract}

\section{Introduction}
\begin{figure*}
    \centering
    \includegraphics[width=\textwidth]{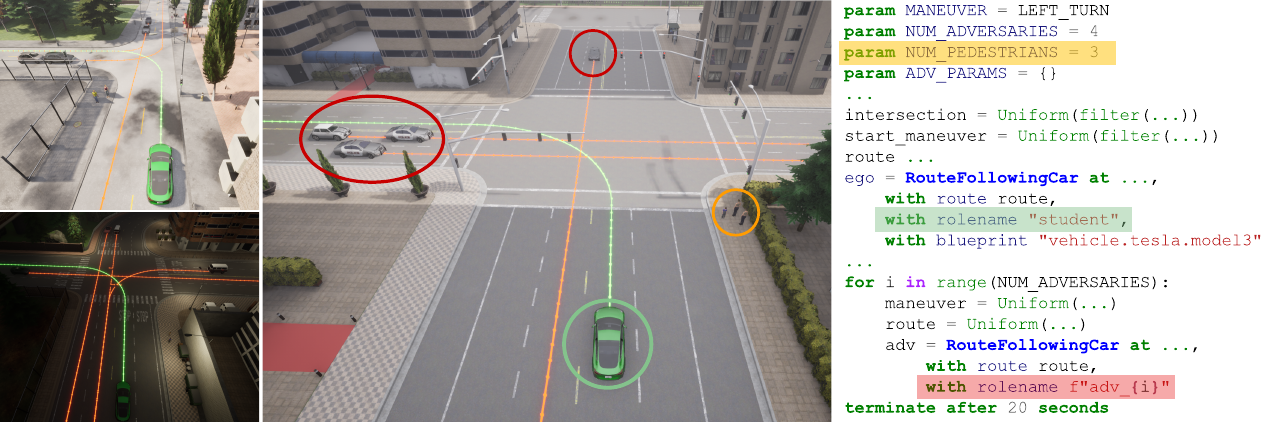}
    \caption{
    This multi-agent scenario illustrates an intersection where five vehicles navigate according to assigned routes, with three pedestrians observed on the sidewalk adjacent to the ego vehicle.
    A visual representation of the simulation is depicted in the center, while the scenario description from which the simulation parameters are sampled is described on the right. On the left, we depict other scenarios sampled from the same Scenic description.}
    \label{fig:scenario-demo}
\end{figure*}

Autonomous driving (AD) is a groundbreaking frontier of transportation technology, offering the promise of safer, and more efficient roadways.
However, full autonomy is still a long-standing, but not-yet achieved goal for researchers across generations. 
As recent advancements enable the deployment of more vehicles in cities worldwide, extensive testing in a diverse range of scenarios becomes essential.
For every kilometer driven by AD vehicles on real-world roads, they traverse several orders of magnitude more in simulation. Thus, access to high-fidelity simulation environments plays a crucial role in advancing the field of AD. 

Open-source simulation software like CARLA~\cite{dosovitskiy_carla_2017} has already empowered researchers and practitioners to evaluate and train AD algorithms in realistic environments. Despite this, creating diverse and lifelike traffic scenarios incorporating varying numbers of agents remains a laborious and time-consuming task.
To tackle this challenge, scenario specification languages like Scenic~\cite{fremont_scenic_2019,Fremont2023} and OpenSCENARIO\footnote{\url{https://www.asam.net/standards/detail/openscenario/}} have been introduced. These scenario specification approaches streamline the process of defining logical traffic scenarios and are well-integrated into simulation environments such as CARLA. However, we observe that: (1) infrastructure aiding the training and evaluation of AD stacks is scattered across numerous ad hoc implementations, (2) integration of scenario specifications and simulation environments lacks support for training in multi-agent systems, and (3) approaches targeting open-ended curriculum generation have limited support.
To address these shortcomings, we propose \adexgym, an open-source multi-agent training framework that leverages existing traffic scenario-specification approaches for CARLA. Additionally, we demonstrate how \adexgym can be used to implement novel auto-curriculum approaches~\cite{jiang_replay-guided_2022} for generating open-ended traffic scenarios with adaptive difficulty. 
The key contributions of \adexgym are:
\begin{enumerate}
    \item A multi-agent, scenario-based AD training and evaluation framework for the CARLA simulator. 
    \item A versatile and flexible configuration infrastructure for our framework that supports multiple scenario description and sampling solutions, sensor models and levels of abstraction for defining action and state spaces.
    \item Experimental evaluation to demonstrate \adexgym's practicality for multi-agent learning and ability to generate auto-curricula from scenario descriptions.
    \item Extensive comparison and classification in the context of AD simulation frameworks.
\end{enumerate}

\subsection{Illustrative Scenario}
In this scenario, as depicted in Figure~\ref{fig:scenario-demo}, five vehicles converge on a four-way intersection, each following a predefined route toward individual goal locations. Additionally, three pedestrians traverse the sidewalk near the intersection. All vehicles are equipped with sensors, including cameras, odometry, and velocity sensors.
Despite the simplicity of this setting, the potential variations are vast, depending on initial conditions such as the number and types of actors and their behaviors. Effective training needs exposure to diverse scenario variations for robust generalization. Moreover, meticulous selection of progressively complex challenges is crucial but requires expert knowledge.
In the following, we introduce MATS-gym and showcase its effectiveness in generating scenarios of varying difficulty.


\section{Related Work}
\begin{table*}
    \caption{Comparison of various traffic scenario frameworks}
    \label{tab:comparison}
    \begin{center}
        \begin{tabular}{|c||c|c|c|c|c|c|c|c|c|}
            \hline
            \multirow{2}{*}{\textbf{Framework}} & \multicolumn{2}{c|}{\textbf{Scenario Specification}} &  \multicolumn{3}{c|}{\textbf{Realism}} & \multicolumn{2}{c|}{\textbf{Traffic Types}} & \multicolumn{2}{c|}{\textbf{Training}} \\
            \cline{2-10}

            & Sampling & Scriptable & Sensors & Visual & Physics & Highway & Urban & RL & MARL \\
            \hline

            HighwayEnv \cite{leurent_environment_2018} & \xmark & $\sim$ & \xmark  &\xmark & \xmark  & \cmark & $\sim$ & \cmark & \cmark \\
            \hline
            BARK \cite{bernhard_bark_2020} & \xmark & \cmark & $\sim$ & \xmark & \xmark & \cmark & $\sim$ & \cmark & \xmark \\
            \hline
            CommonRoads \cite{althoff_commonroad_2017,wang_commonroad-rl_2021} & \xmark & \cmark &  \xmark & \xmark & \cmark & \cmark & $\sim$ & \cmark & \xmark \\
            \hline
             SMARTS \cite{zhou_smarts_2020} & \cmark & \cmark & $\sim$ & \xmark & \cmark & \cmark & $\sim$ & \cmark & \cmark \\
            \hline
             MetaDrive \cite{li2021metadrive} & \cmark & $\sim$ & \cmark & \xmark & $\sim$ & \cmark & \cmark & \cmark & \cmark \\
            \hline
             DI Drive \cite{contributors_di-drive_2021} & \xmark & \cmark & \cmark & \cmark & \cmark & \cmark & \cmark & \cmark & \xmark \\
            \hline
            MACAD-Gym \cite{palanisamy_multi-agent_2020} & \xmark & $\sim$ & \cmark & \cmark & \cmark & \cmark & \cmark & \cmark & \cmark\\
            \hline
           \hline
            \adexgym (ours) & \cmark & \cmark & \cmark & \cmark & \cmark & \cmark & \cmark & \cmark & \cmark \\
            \hline
        \end{tabular}
    \end{center}
\end{table*}

\par{\textbf{Training Frameworks for AD.}}
We compare our approach to other frameworks commonly utilized for training and evaluating AD agents. We identified several popular frameworks and assess them based on several key aspects crucial for our use-case. These aspects include ease and expressivity of scenario specifications, simulation realism, supported traffic types, and integrated training interfaces. Table \ref{tab:comparison} summarizes the comparison.

HighwayEnv~\cite{leurent_environment_2018}, BARK~\cite{bernhard_bark_2020}, SMARTS \cite{zhou_smarts_2020}, and the CommonRoads Suite~\cite{althoff_commonroad_2017,wang_commonroad-rl_2021} offer alternatives to full-fledged game-engine simulations like CARLA. They focus on modeling vehicle dynamics and motion planning, with CommonRoads providing more sophisticated multi-body dynamics compared to HighwayEnv and BARK. While simpler environments are computationally efficient, they lack complex sensor models and realistic observations. These frameworks mainly involve vehicles and lack pedestrian or object modeling found in urban traffic. BARK and CommonRoads support scriptable, and HighwayEnv custom scenarios through code re-implementation, respectively. SMARTS allows to model urban traffic with traffic lights and traffic signs. However, as the focus is not on graphical realism, those scenarios can not reach the visual diversity of traffic scenes in CARLA, which is necessary to train and evaluate vision based AD stacks. 

MetaDrive~\cite{li2021metadrive} is a training framework that builds on the Panda3D engine. It offers a wide range of features such as procedural road generation and it incorporates scenarios from different traffic datasets. However, scenarios can not be specified by behavioral building blocks or by predefined sub-scenarios. Moreover, its engine limits the structural variety of environments that can be achieved in CARLA.

DI Drive Gym~\cite{contributors_di-drive_2021} and MACAD Gym~\cite{palanisamy_multi-agent_2020} are both training frameworks designed around CARLA. DI Drive Gym focuses on single-agent autonomous driving systems and interfaces with ScenarioRunner scenarios, lacking support for training multi-agent systems or scenario generation via sampling. MACAD Gym supports multi-agent training but offers limited scenario generation capabilities, allowing only manual specification of initial conditions.

\adexgym in contrast, supports scenario scripting and sampling, realistic sensors, physics, and rendering, urban as well as highway traffic, and (multi-agent) reinforcement learning. Our comparison focuses on relevant aspects for learning-based approaches in realistic urban traffic scenarios. We acknowledge that our criteria may not be complete, and there might be aspects that could be relevant in other use cases. However, we tried to cover a broad spectrum of frameworks that could be used for training and evaluation.

\par{\textbf{Curriculum Learning.}}
Generating safety-critical scenarios is an active area of research and has been addressed in numerous works in recent years~\cite{wenhao_2023_adversarial}.
However, training agents in complex environments requires presenting scenarios at the \emph{right difficulty}, matching their capabilities to generate meaningful learning signals. Curriculum Learning (CL) and Auto CL (ACL) methods have emerged to progressively expose agents to more complex environments. While traditional CL involves manually designing curricula, ACL automates this process by dynamically presenting suitable scenarios based on agent performance. For an overview, we refer to~\cite{portelas_2021_curriculum}.

Previous works also applied ACL to AD scenarios~\cite{qiao_2018_curriculum_ad,anzalone2021curriculum,anzalone_2022_curriculum}.
In this work, we adopt \emph{Unsupervised Environment Design} for AD, a recent variation of the ACL methodology where task creation itself is treated as an optimization problem, allowing for more flexible scenario adaptions.

\section{Background}
We model the environment as a \textbf{Partially Observable Stochastic Game (POSG)}, defined as a tuple:
\[
\mathcal{G} = \langle \mathcal{N}, \mathcal{S}, \{\mathcal{A}_i\}_{i \in \mathcal{N}}, \{\mathcal{O}_i\}_{i \in \mathcal{N}}, \mathcal{T}, \mathcal{R}, \mathcal{Z}, \gamma \rangle.
\]

It consists of the set of agents \( \mathcal{N} = \{1, 2, \dots, N\} \) and the set of environment states \( \mathcal{S} \). For each agent \( i \in \mathcal{N} \), \( \mathcal{A}_i \) denotes the set of actions, and \( \mathcal{O}_i \) is the set of observations. The joint action space is defined as \( \mathcal{A} = \mathcal{A}_1 \times \mathcal{A}_2 \times \cdots \times \mathcal{A}_N \), the joint observation space is \( \mathcal{O} = \mathcal{O}_1 \times \mathcal{O}_2 \times \cdots \times \mathcal{O}_N \).
The state dynamics are given by \( \mathcal{T}: \mathcal{S} \times \mathcal{A} \rightarrow \Delta(\mathcal{S})\), where \( \mathcal{T}(s' | s, \mathbf{a}) \) represents the probability of transitioning from state \( s \) to state \( s' \) given the joint action \( \mathbf{a} \in \mathcal{A} \). The reward function \( \mathcal{R}: \mathcal{S} \times \mathcal{A} \rightarrow \mathbb{R}^N \) specifies the rewards for each agent, where \( \mathcal{R}(s, \mathbf{a}) = (R_1(s, \mathbf{a}), R_2(s, \mathbf{a}), \dots, R_N(s, \mathbf{a})) \).
The observation function \( \mathcal{Z}: \mathcal{S} \times \mathcal{A} \rightarrow \Delta(\mathcal{O}) \) gives the probability \( \mathcal{Z}(\mathbf{o} | s, \mathbf{a}) \) of receiving joint observation \( \mathbf{o} \in \mathcal{O} \) in state \( s \) following joint action \( \mathbf{a} \). The discount factor \( \gamma \in [0, 1] \) represents the relative importance of future rewards.
At each time step, the system is in a state \( s \in \mathcal{S} \), which is not directly observable. Each agent selects an action \( a_i \in \mathcal{A}_i \) based on their observation \( o_i \in \mathcal{O}_i \) by following a stochastic policy $\pi_i: \mathcal O_i \mapsto \Delta(\mathcal A_i)$. The joint action \( \mathbf{a} = (a_1, a_2, \dots, a_N) \) is applied, resulting in a new state \( s' \in \mathcal{S} \) according to the transition function \( \mathcal{T} \), and each agent receives a reward \( {R_i}(s, \mathbf{a}) \) and a new observation according to \( \mathcal{Z} \).
The value function for a policy $\pi$ is the expected sum of discounted future rewards:
$$
V_i^\pi(s) = \mathbb{E}_{\mathcal T}\left[\sum_{t=0}^{H} \gamma^t R_i(s_t, \mathbf{a}_t) \mid s_0 = s, a^i \sim \pi_i \right].
$$
The goal for agent $i$ is to find a policy $\pi$ that maximizes $V_i^\pi(s)$ over the horizon $H$. For simplicity, we will drop the agent indices in the remainder of the paper.

We allow that the POSG can be modified at runtime by parametrizing the environment dynamics and the set of agents.
This can be formalized as a \textbf{Underspecified POSG} (analogous to \cite{dennis_emergent_2021}):
\[
\mathcal{G}^\Theta = \langle \mathcal{N}^\Theta, \mathcal{S}, \{\mathcal{A}_i\}_{i \in \mathcal{N}}, \{\mathcal{O}_i\}_{i \in \mathcal{N}}, \mathcal{T}^\Theta, \mathcal{R}, \mathcal{Z}, \gamma, \Theta \rangle,
\]
where $\Theta$ is a set of parameters that can be chosen by the scenario generator $\tilde{\pi} \in \Delta(\Theta)$. Unsupervised Environment Design (UED) algorithms seek to find  generators that maximize the regret of a policy $\pi$:
\[
\max_{\tilde{\pi}}  \mathbb{E}_{\theta \sim \tilde{\pi}} \big[ \max_{\pi'} V^{\pi'} - V^\pi \big],
\]
the difference of the maximum achievable return and the actual return. This yields a suitable objective for curriculum design: low regret scenarios indicate that the policy already performs well, while high regret entails difficult but \emph{solvable} scenarios.
Recent works have explored different algorithms for UED. Prioritized Level Replay (PLR) utilizes domain randomization (DR) to generate parameters and a replay buffer to resample levels proportional to their regret, a measure of suboptimality~\cite{jiang_prioritized_2021}. PAIRED~\cite{dennis_emergent_2021} optimizes the level generator via reinforcement learning (RL) to maximize regret, while REPAIRED~\cite{jiang_replay-guided_2022} combines PLR with the level generation approach from PAIRED in a Dual Curriculum Design (DCD) framework. Other approaches use evolutionary algorithms to optimize the generator and include a level editor for environment modifications~\cite{parker-holder_evolving_2022,mediratta_stabilizing_2023}. 
In contrast, \adexgym takes advantage of the Scenic's parametrizable scenario specifications, as a way to build an adaptive scenario generation procedure, and adopts a dual-curriculum design algorithm similar to REPAIRED~\cite{jiang_replay-guided_2022}.

\section{\adexgym}

\adexgym is a multi-agent training and evaluation framework that allows to generate diverse traffic scenarios in CARLA, a high-fidelity traffic simulator.
The framework is designed to reconcile scenario execution engines, such as Scenic~\cite{fremont_scenic_2019} and ScenarioRunner~\cite{scenario_runner}, with multi-agent training frameworks. It offers comprehensive infrastructure for \textit{agent-state retrieval}, providing ground truth information such as position, velocity vectors, and key traffic events like violations and collisions. 
\adexgym supports a versatile infrastructure for \textit{configuring sensor-suites} for agents, akin to the CARLA AD Challenge, which supports complex sensors, such as cameras, LiDAR, and Radar. \adexgym extends the CARLA BirdEyeView~\cite{martyniak_carla-birdeye-view_2020} framework for bird's-eye view observations, by constructing occupancy gridmaps around agents encoding lane markings, traffic lights, signs, and other road users. 
The agents can also access road-network information and vector-based representations of the map.

We account for \textit{different types of tasks}, by providing action spaces at various levels of abstraction. 
On the lowest level, agents can issue throttle, breaking and steering commands at high frequencies. 
On a higher level, agents can instead provide a target waypoint which is tracked by a  PID controller for smooth driving trajectories. 
Finally, we also provide a discrete action space for high level driving commands, suitable for route planning, behavior prediction, or multi-agent interaction. 
Furthermore, \adexgym allows for defining custom tasks and offers pre-defined tasks related to autonomous driving, including route following, infraction avoidance, and driving comfort, crucial for designing reward functions in reinforcement learning approaches.

\subsection{Scenario Based Curriculum Generation}
\label{sec:dcd}
In the following, we will outline our approach to use Scenic's parametrizable scenario specifications as a way to build an adaptive scenario generation procedure.
We adopt a dual-curriculum design algorithm based on REPAIRED~\cite{jiang_replay-guided_2022}, a recent algorithm combining PLR with an adaptive scenario generator. 
Algorithm~\ref{alg:mats-repaired} outlines this approach.
At each iteration, we either generate a new scenario by sampling environment parameters from generator $\tilde{\pi}$ or we sample a replay scenario from the PLR buffer.
The buffer samples scenarios with a probability proportional to their regret and a staleness metric, as described in the original approach.
Then, a trajectory is sampled from the environment to update the policy $\pi_\phi$ parametrized by $\phi$ (e.g. a neural network) using any RL algorithm (e.g. PPO \cite{schulman2017proximal}). In the next step, we compute the estimated regret for the current scenario to update the buffer and the environment generator.
Differently than the original, we do not use an antagonist agent to estimate the regret.
Instead, we use the \emph{Maximum Monte Carlo}~\cite{jiang_replay-guided_2022} regret estimator which approximates the regret to avoid the expensive inner optimization routine:
\begin{equation}
\label{eq:regret}
    \textsc{Regret}(\pi) = \max_{\pi'} V^{\pi'} - V^\pi \approx \frac{1}{H} \sum_{t=0}^H R_\text{max} - V^\pi(s_t).
\end{equation}
This regret formulation compares the maximum achieved return for a scenario with the expectation of the returns for the current policy.
Furthermore, for the scenario generator optimization, we use an off-the shelf black-box optimization algorithm~\cite{akiba2019optuna}.

\begin{algorithm}
\caption{MATS-REPAIRED}
\label{alg:mats-repaired}
\begin{algorithmic}[1]

\STATE {\bfseries Input:} policy $\pi_\phi$, scenario generator $\tilde{\pi}$, buffer $\Lambda$
\WHILE{not converged}
    \STATE Sample replay decision $d \sim P_{D}(d)$
    \IF{d=0}
        \STATE Generate scenario parameters: $\theta \sim \tilde{\pi}$
        \STATE Insert $\theta$ into buffer $\Lambda$. 
    \ELSE
        \STATE Sample replay scenario: $\theta \sim \Lambda$
    \ENDIF
    \STATE Sample batch $\tau$ with $\theta$: $\{(o_i,a_i,r_i,o'_i)\}^N_{i=0} \sim \mathcal{T}_{\pi_\phi}^\theta$
    \STATE Policy gradient step: $\phi \leftarrow \phi + \alpha \, \nabla_\phi J^{\pi_\phi}_{\textsc{PPO-Loss}}(\tau)$
    \STATE Estimate regret $\hat{R} \approx \textsc{Regret}(\pi)$ using Eq.~\ref{eq:regret}
    \STATE Update scenario buffer: $\Lambda \leftarrow (\theta, \hat{R})$
    \IF{scenario was generated}
        \STATE $\tilde{\pi} \leftarrow \textsc{Black-Box-Opt}(\theta, \hat{R})$
    \ENDIF
\ENDWHILE
\RETURN $\pi_\phi$
\end{algorithmic}
\end{algorithm}

\section{Experiments}

We conduct two experiments in which we show that \adexgym can be used to: (1) Train multiple agents in an implicit coordination task, and (2) For auto-curriculum generation.
Both experiments share the base scenario in which agents are tasked with successfully navigating along a route through an intersection in urban environments, similar to the one introduced in the illustrative example.

The reward function for this task is a combination of a progress-based and a cruise speed reward:
\begin{equation}
r_t = r_\text{progress} + r_\text{cruise},
\end{equation}
where $r_\text{progress} = p_t - p_{t-1}$ is the progress made along the route since the last time-step and $r_\text{cruise} = min(\frac{v_t}{v_\text{target}}, 1)$  rewards the agent to drive at a cruise velocity $v_{target}$.
For the optimization of the agent policy we use Proximal Policy Optimization (PPO)~\cite{schulman2017proximal}.
The agent receives birdview observations as input and can therefore observe the drivable road area, lane markings, route and other traffic participants.

\subsection{Learning in Different Action Spaces}
\begin{figure*}[h!]
    \centering
    \includegraphics[width=\textwidth]{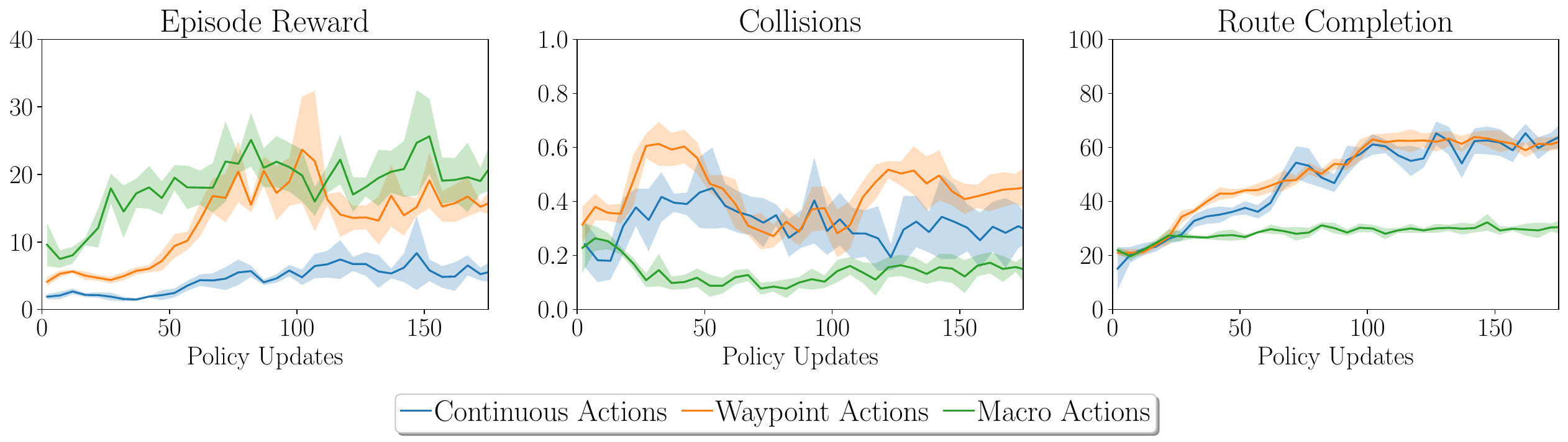}
    \caption{
    Learning curves for I-PPO under different action definitions and the impact on 
    episodic return, collisions and route completion.
    Performance reports mean and standard deviation over 5 consecutive policy updates
    of the same run.
    }
    \label{fig:experiment-actions}
\end{figure*}
In this experiment, we demonstrate the framework's usability for multi-agent training and discuss how the choice of actions affects the training of cooperative agents.

\par{\bf Scenario Description.}
Based on the initial scenario, we let four vehicles navigate through a four-way intersection, each with a designated route and a time limit to reach their destination. The starting conditions and goals vary per episode and are sampled from the description. To succeed, the agent must manage its vehicle, collaborate with other traffic participants, and follow the prescribed route. Termination occurs when all agents are stopped.

\par{\bf Training Setup.}
Each vehicle is governed by an independent policy,
which perceives a birdview observation encoding the road layout,
vehicle positions relative to the agent, and route information.
To study the impact of actions spaces,
we consider the same task definition
with \textit{continuous}, \textit{waypoint} and \textit{macro} actions.
Each of the proposed actions have different control frequencies:
continuous actions are repeated twice, 
waypoint actions are executed for 5 steps, 
and macro actions persist for 10 steps with 0.05s per step.
We train the policies with Independent PPO~\cite{yu2022surprising} and account for the different action frequencies by fixing the training budget to $175$ policy updates, each over a batch of $2048$ transitions.

{\bf Results.}
Figure \ref{fig:experiment-actions} shows the learning curves with respect to average episodic reward, collisions and route completions, evaluated over the batch of training data for each policy update.
We aggregate the metrics over $k$-sized bins of policy updates ($k=5$), 
reporting the mean and standard deviation.

We observe different characteristics of the learned policies. 
Low-level and waypoint target actions incur a high number of collisions, especially in early training stages. On the other hand, macro actions make use of a basic lane keeping and collision avoidance controller, which prevents agents from leaving drivable areas and avoids most of the accidents, leading to a low collision rate throughout the training.
This restriction comes at the cost of frequent deadlocks, which result in a low route completion rate. 
Less restrictive action spaces allow agents to leave the predefined lane to conduct evasive maneuvers, leading to higher average route completion in later training stages.

This experiment emphasizes how the action space design profoundly shapes emergent behavior and affects the learning task's difficulty. Careful modelling of the problem, encompassing observations and actions, is crucial for multi-agent learning and serves as a key feature in training frameworks. Importantly, it demonstrates that there is no universally optimal level of abstraction for the action space; instead, the choice depends on the problem at hand.

\subsection{Scenario Based Environment Design}
\begin{figure*}[h!]
    \centering
    \includegraphics[width=0.9\textwidth]{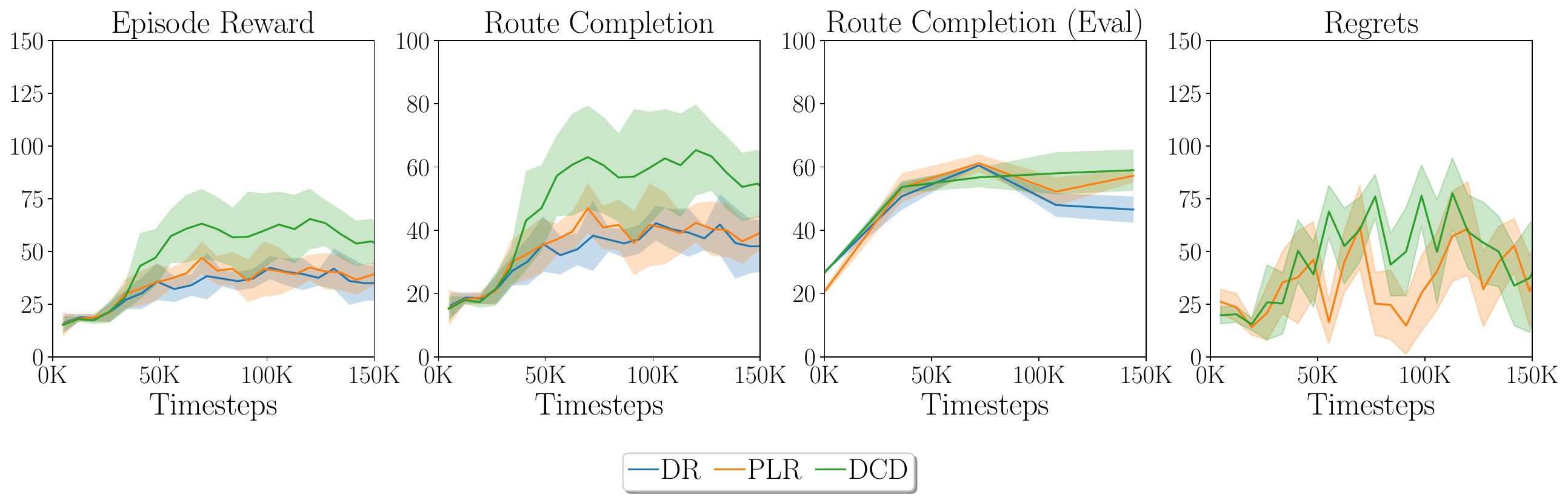}
    \caption{
    Learning curves of average episodic return, route completion during training and evaluation over 3 seeds with one standard deviation. We also report the average regret of the level buffers of PLR and our DCD approach over timesteps.
    }
    \label{fig:ued_results}
\end{figure*}
In this experiment, we assess the ability of the curriculum design algorithm to align the training scenario distribution with the agent's capabilities. We examine its impact on agent performance and the distribution of generated scenarios. Additionally, we compare its effectiveness with PLR and basic domain randomization. We analyze the evolution of sampled scenario parameters for each approach and evaluate their adaptation to the agent's capabilities, particularly in the early stages of training when the agent struggles with complex scenarios.

\par{\bf Scenario Description.}
We parameterize the base scenario introduced before 
with a variety of discrete and continuous variables,
which characterize the difficulty of the generated scenes.
Intuitively, learning to follow a straight path through an empty intersection is easier than performing an unprotected left turn in a busy intersection where other traffic participants are driving recklessly.
In our experiment, we define the following parameters that can be sampled:
\begin{itemize}
    \item Assigned route (straight, left or right turn).
    \item Number of other vehicles in the intersection.
    \item Target speed for the other agents.
    \item Whether the other vehicles keep safety distances.
    \item Whether the other vehicles respect traffic lights.
\end{itemize}

{\bf Results.}
\begin{figure*}
    \centering
    \includegraphics[width=\textwidth]{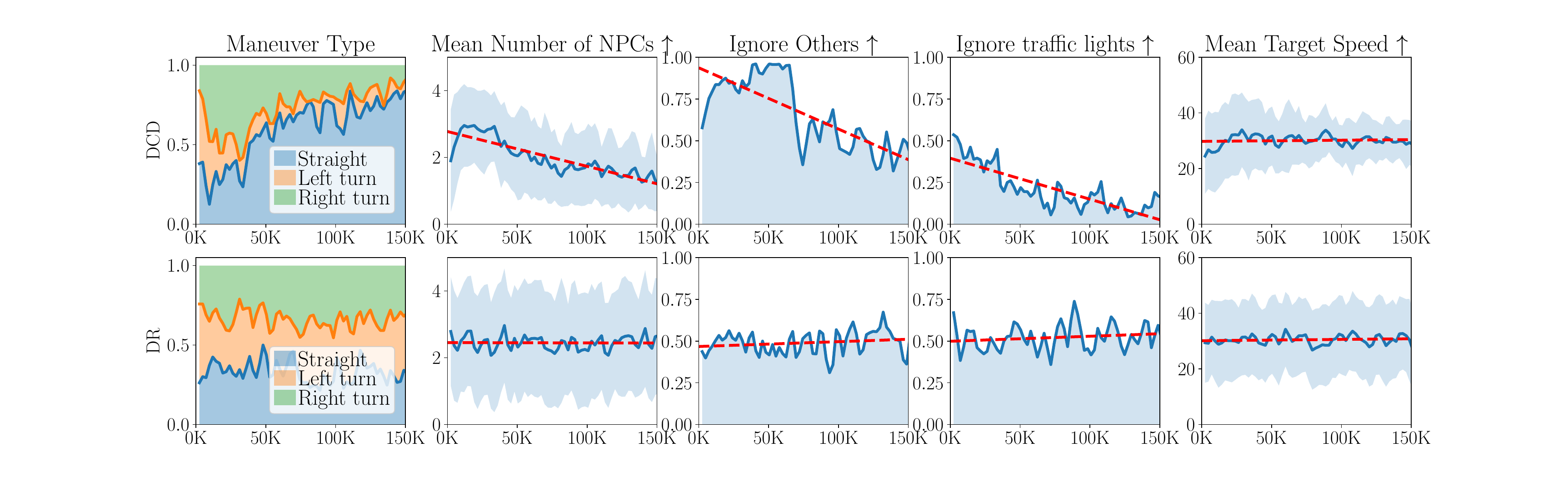}
    \caption{
    Evolution of the parameter distribution during the training with DCD (top) and domain randomization (bottom). The plot shows the fractions of route types, mean number of NPCs, the fraction of agents that do not keep safety distances and ignore traffic lights and the mean target speed.
    For each parameter,     
    we report mean and standard deviation over the batch of training data and indicate the direction of major difficulty ($\uparrow$).
    We observe that DCD steers the parameter distribution towards configurations that are supposedly easier to solve: straight crossings, fewer and safer NPCs, etc. This leads us to the conclusion that DCD adapts the training distribution faster to the performance level of the agent.
    }
    \label{fig:ued_params}
\end{figure*}
\begin{figure*}[h!]
     \centering
     \includegraphics[width=0.7\textwidth]{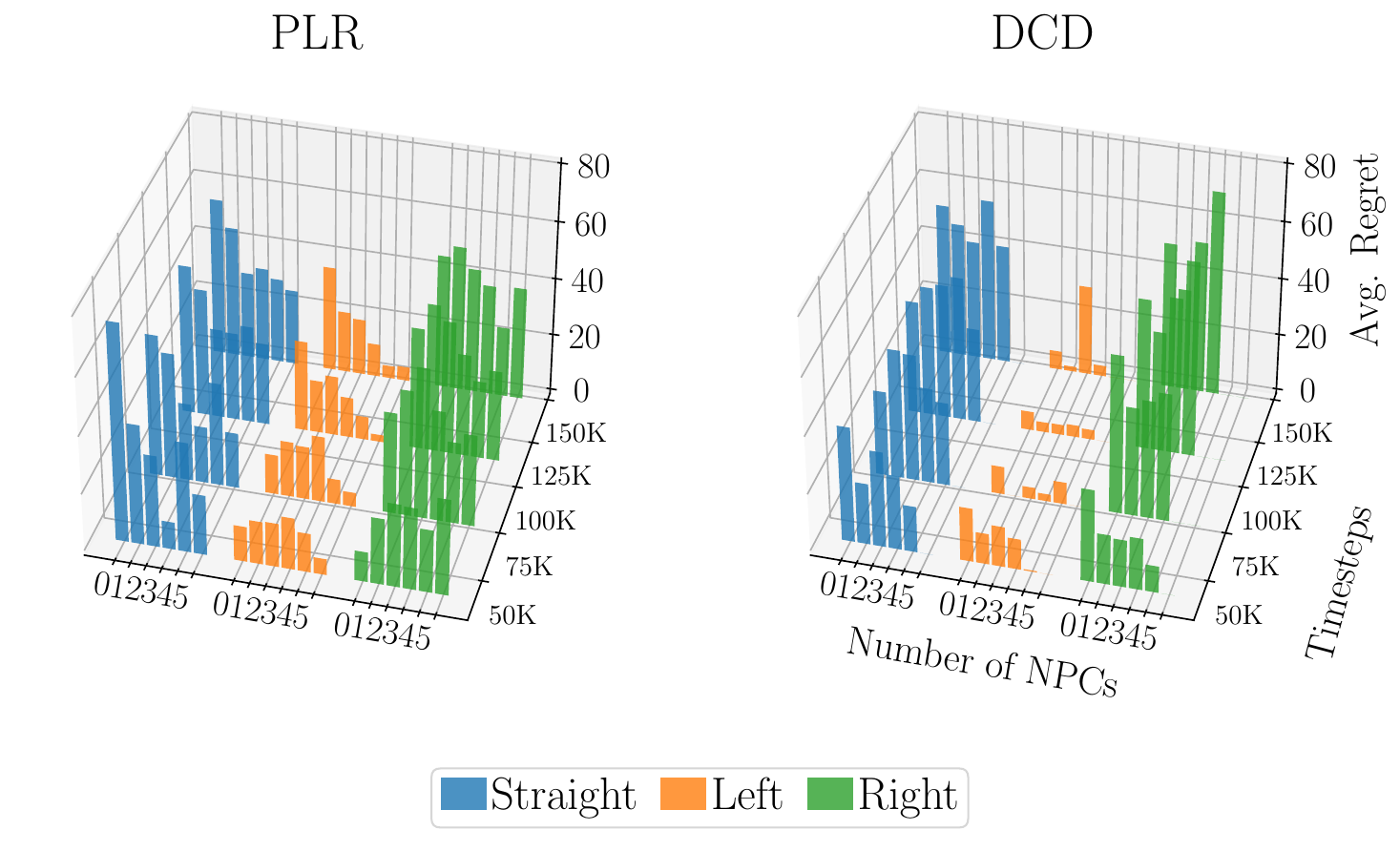}
        \label{fig:Scenario regret per parameter combination}
     \hfill
    \caption{For each pair of maneuver type  and number of NPCs, we average the regrets of all scenarios with the same parameters in the buffer at four checkpoints throughout the training. We observe that the environment generation policy of DCD leads to more narrowly concentrated regret on fewer configurations, suggesting faster adaption of the scenario sampling.}
    \label{fig:level_stats}
    \vspace*{-3ex}
\end{figure*}
In Figure~\ref{fig:ued_results}, we compare learning curves over 150K environment steps. Both PLR and DR show similar performance in terms of episodic returns and route completion during training, indicating comparable scenario difficulty. DCD demonstrates a notable increase in training signal, suggesting adaptation by generating progressively easier scenarios in the initial stages of training. We evaluate all approaches on the same 12 hold-out scenarios, representing different maneuvers with varying numbers of NPCs.
Although the performance on the hold-out set is comparable in early stages of the training, DCD and PLR do not suffer as much from performance drops in later stages.

Figure~\ref{fig:ued_params} provides insight into the progression of scenario parameters throughout training. DCD shifts the scenario distribution towards less challenging scenarios, characterized by straight maneuvers for the agent and a low number of vehicles in intersections. For comparison, we also show how DR naturally maintains a uniform distribution over environment parameters.

We also investigate the distribution of scenario parameters in the replay buffers of PLR and REPAIRED. 
Figure~\ref{fig:level_stats} depicts the average regret associated with various parameter combinations at four distinct checkpoints during the training process. For visualization purposes, we focus on combinations involving the maneuver type and the number of other vehicles (NPCs) present in the intersection.
Observing the data, it becomes evident that the average regrets of Dual-Curriculum Design (DCD) scenarios tend to be higher and exhibit a narrower distribution across fewer parameter combinations compared to those of PLR alone. This trend can be attributed to the presence of the adaptive scenario sampler, which facilitates expedited convergence towards parameter combinations that are more pertinent to the task at hand.
Our experiments demonstrate the effectiveness of automatic curriculum design in aligning scenario generation with the agent’s capabilities. Additionally, optimizing the scenario generation process, rather than just the resampling procedure, accelerates the adaptation of the scenario distribution.

\section{Conclusion}
In this work,  we present \adexgym, a multi-agent training framework capable of generating scenario-based auto-curricula for AD tasks in CARLA. Leveraging Scenic allows us to sample from scenario distributions and enables the integration with UED approaches. 
This compatibility resembles a promising way to generate more relevant and realistic training scenarios for AD.
We demonstrate the usability of our framework 
in two experiments of multi-agent training and automatic curriculum generation, respectively.
By introducing \adexgym and demonstrating its application in various experiments, 
we contribute to ongoing efforts in advancing multi-agent training for autonomous driving.

\section{acknowledgements}
This project has received funding from 
the Austrian FFG ICT
of the Future program under grant agreement No 880811.



\bibliographystyle{IEEEtran}
\bibliography{references}
\end{document}